\pdfoutput=1
\documentclass[fleqn]{article}
\makeatletter
  \newcommand\figcaption{\def\@captype{figure}\caption}
  \newcommand\tabcaption{\def\@captype{table}\caption}
\makeatother

\PassOptionsToPackage{numbers, compress}{natbib}
\usepackage[final]{nips_2016}

\usepackage[utf8]{inputenc} % allow utf-8 input
\usepackage[T1]{fontenc}    % use 8-bit T1 fonts
\usepackage{hyperref}       % hyperlinks
\usepackage{url}            % simple URL typesetting
\usepackage{booktabs}       % professional-quality tables
\usepackage{amsfonts}       % blackboard math symbols
\usepackage{nicefrac}       % compact symbols for 1/2, etc.
\usepackage{microtype}      % microtypography

\usepackage{amsmath}
\usepackage{bm}
\usepackage{graphicx,epstopdf}
\usepackage{enumitem}
\usepackage{color}

\usepackage{multirow}
\usepackage{subtable}
\usepackage{minipage}
\usepackage{caption}

\title{Storytelling of Photo Stream with Bidirectional Multi-thread Recurrent Neural Network}

\author{
  Yu Liu \\
  University at Buffalo, SUNY\\
  \texttt{yliu44@buffalo.edu} \\
  \And
  Jianlong Fu\\
  Microsoft Research Asia\\
  \texttt{jianf@microsoft.com}\\
  \And
  Tao Mei\\
  Microsoft Research Asia\\
  \texttt{tmei@microsoft.com}\\
  \And
  Chang Wen Chen\\
  University at Buffalo, SUNY\\
  \texttt{chencw@buffalo.edu}\\
}

\begin{document}

\maketitle

\begin{abstract}
  Visual storytelling aims to generate human-level narrative language (i.e., a natural paragraph with multiple sentences) from a photo streams. A typical photo story consists of a global timeline with multi-thread local storylines, where each storyline occurs in one different scene. Such complex structure leads to large content gaps at scene transitions between consecutive photos. Most existing image/video captioning methods can only achieve limited performance, because the units in traditional recurrent neural networks (RNN) tend to ``forget'' the previous state when the visual sequence is inconsistent. In this paper, we propose a novel visual storytelling approach with \emph{Bidirectional Multi-thread Recurrent Neural Network} (BMRNN). First, based on the mined local storylines, a \emph{skip gated recurrent unit} (sGRU) with delay control is proposed to maintain longer range visual information. Second, by using sGRU as basic units, the BMRNN is trained to align the local storylines into the global sequential timeline. Third, a new training scheme with a storyline-constrained objective function is proposed by jointly considering both global and local matches. Experiments on three standard storytelling datasets show that the BMRNN model outperforms the state-of-the-art methods. %with $xx\%, xx\%, xx\%$ on NYC, Disney and SIND datasets, respectively, in terms of recall@1 in story retrieval tasks.
\end{abstract}

\section{Intruduction}

 After the tremendous success recently achieved in image captioning and video captioning tasks \cite{OrderEmbeding,Unifying,CVPR2015Karpathy,CVPR2015Vinyals,ICLR2015Mao,ICCV2015Venu,CVPR2015Donahue,NAACL2015Venu}, it becomes promising to move forward to resolving \emph{visual stories}, sequences of images that depicts events as they occur and change \cite{VisualStorytelling}. In this paper, we focus on \emph{Visual Storytelling}, a new vision-to-language task that translates images streams to sequences of sentences as stories.

 However, it is a challenging issue due to the visual story's complex structure, which we call \emph{cross-skipping structure}. To our observation, a typical story usually has a global sequential timeline and multi-thread local storylines, analogy to a movie of story always narrates along a global time axis while crosscutting between multiple sub-stories. For example, figure 1 shows a typical visual story, with black arrows as timeline and blue/red arrows as two cross-skipping storylines. Intuitively, photos with similar scenes are likely to share a storyline, and thus the cross-skipping structure can be automatically detected by considering scene similarity (Details in Section 3.1). We emphasize that cross-skipping is typical phenomenon in majority of visual stories, taking a 74.9\% partition of a public visual story dataset \cite{VisualStorytelling} in our measure. However, this cross-skipping structure is a disaster for current exist sequential models, such as RNN with GRU or LSTM \cite{ICCV2015Venu,CVPR2015Donahue,NAACL2015Venu}. It can lead to very large gaps at the boundary between skips, which contains long-term dependencies that RNN cannot practically learn due to its limited memory on long intervals in input/output sequence \cite{LongTerm}.

 However, in our case, this particular type of gap can be bridged, as we call it \emph{pseudo-gap}, by using the skip information in each storyline. In the visual story with five photos of figure 1, for example, the fourth photo is largely gapped from the third in visual. But to model the fourth, we bridged the gap by also using the information of the first photo. To this end, we propose a novel neural network called \emph{skip Gated Recurrent Unit} (sGRU) (Section 2), and a \emph{Bidirectional Multi-thread RNN} (BMRNN) architecture (Section 3), to solve the pseudo-gap problem. Specifically, rather than a normal RNN model can only model global timeline, our BMRNN is designed also to capture the skip information along storylines. And we equip sGRU with a preservation scheme that is able to memorize the captured skip information and compensate the pseudo-gap. Seamlessly integration of the two design by using the sGRU as basic units in BMRNN is proposed to solve this problem.

%\emph{delete the following paragraph here, you can copy to section 3 as a summary of the flowchart.}
%Figure 1 shows the framework of our approach. We will first construct an embedding space using individual image-sentence pairs to better represent both visual and textual signals into a same space. Then, in the embedding space, we build a bidirectional architecture using the proposed Multi-thread RNN to interpret the visual stream into sentences sequences considering its complex visual structure. The output sentence sequence should have the consistent structure that each of them is coherent with others both in global timeline and local storylines, just as the input visual signals. Finally, a single-layer RNN is employed as a language model to provide a further smoothness on the textual domain to produce fluent content.

\begin{figure}[t]
\centering
   \includegraphics[width = 1.0\linewidth]{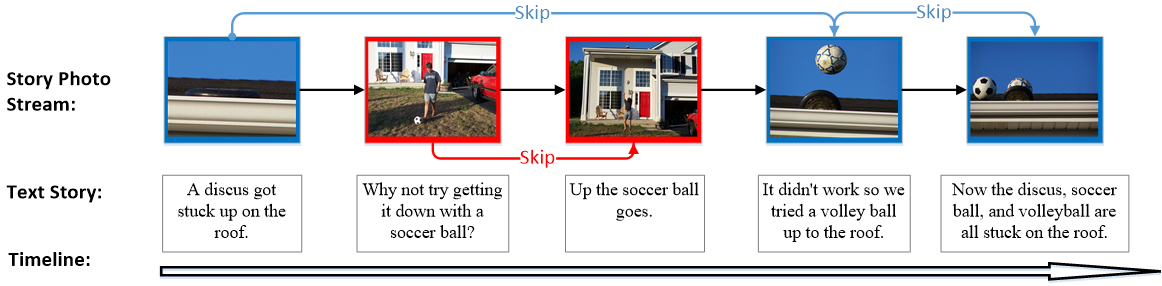}
   \caption{\small The skip-structure of a visual story example. Intuitively, the photos that share the same scene can form local \emph{skips} along timeline. Here we have two skips denoted by blue and red arrows. The black arrow represents the visual flow along global timeline. [Best viewed in color]}
   \label{fig:exe}
\end{figure}
\vspace{-3mm}

To our best knowledge, this work is the first to model visual story structure and generate human-level narrative language. We highlight the contributions as follows:
\begin{itemize}[nosep]
\item We propose a new skip Gated Recurrent Unit (sGRU) as basic units in RNN, with a preservation scheme and non-linear mapping to keep the long-range information that could be lost by general GRU.
\item We built a Bidirectional Multi-thread RNN (BMRNN) framework, which can capture the skip information in visual stories to model the local storyline. A storyline-constrained objective function is further proposed to jointly consider both the global and local matches.
\item We conduce both quantitative and subjective experiments on three widely-used storytelling datasets, NYC, Disney \cite{Kim} and SIND \cite{VisualStorytelling}, and obtain the performance gain of recall@1 over the-state-of-the-art with $22.8\%, 15.6\%, 15.7\%$, respectively.
\end{itemize}

\subsubsection*{Related Works}

Thank to the prospective growing of research interest in translating visual type information into language, there are great number of works raised in vision-to-language modeling. They can be divided into three families by the input/output form: single-frame to single-sentence, multi-frame to single-sentence and multi-frame to multi-sentence.

\textbf{Single-frame to single-sentence modeling} These models focus on image captioning task, which can be classified into two categories: semantic element detection based methods \cite{TPAMI2013Kulkarni,ECCV2010Farhadi,EACL2012Mitchell,EMNPL2011Yang} or Convolutional Neural Network (CNN) based methods\cite{OrderEmbeding,Unifying,CVPR2015Karpathy,CVPR2015Vinyals,ICLR2015Mao}. In semantic element detection based methods, the regions of interest are first detected and represented with low level features. The detected regions are then projected to an intermediate space defined by a group of semantic elements, such as object, action, scene, attributes etc., which then fill in a sentence template. Although this type of visual modeling is intuitive for visual component classification, it cannot contain rich semantics since each visual component requires explicit human annotation. In CNN based models, the fully-connected (FC) layer output of a CNN is extracted to represent the input images for classification \cite{CNN}. It has convincingly shown the ability in image representation, which encourage the surge of recent research in capturing the semantic relation between vision and descriptive language. Hence, it is natural for us to use CNN features as our image representor.

\textbf{Multi-frames to single-sentence modeling} This family of models, mainly focus on video captioning, captures the temporal dynamics in variable-length of video frames sequence, to map to a variable-length of words \cite{ICCV2015Venu,CVPR2015Donahue,NAACL2015Venu}. The sequence-to-sequence modeling are mainly relied on a RNN framework, such as Long-Short-Term-Memory (LSTM) networks \cite{LSTM}. Moreover, bidirectional RNN (BRNN) is explored recently to model the sequence in both forward and backward pass \cite{VideoBRNN}. However, the shortcoming of all above works is that, within one model, they can only capture temporal structure of a short video clip with near-duplicated consecutive frames, due to the design of normal RNN unit (details in Section 2.2) that cannot learn long-dependencies at large intervals in sequences \cite{LongTerm}.

Similar to our problem formulation, \cite{ICCV2015Yao} argues that a video has local-global temporal structure, where the local structure refers to the fine-grained motion of actions. They employ a 3D CNN to extract local action feature and an attention-based LSTM to exploit the global structure. However, the long-dependency problem is still unsolved that their model can only handle the local structures of video segments consisting of consecutive near-duplicated frames.

\textbf{Multi-frame to multi-sentence modeling} \cite{Kim} claims to be the first to handle the image streams to sequences of sentences task. They use a coherence model in textual domain, which resolves the entity transition patterns frequently appear in the whole text. A bidirectional RNN is used to model the sentence sequence to match the images. However, they mainly focus on textual domain coherence rather than visual domain structure. In contrast, we believe that interpretation from vision domain is more confirm to human-cognition-way for vision-to-language purposed task, and should be more suitable for the visual storytelling task. We validate this point in our experiments in Section 4.2.

 %compare us with their work in task of visual storytelling and shows superior results in experimental evaluation.

\section{Skip Gated Recurrent Unit (sGRU)}

In this section we introduce our designed sGRU model. We first review the Gated Recurrent Unit \cite{GRU}. Then we introduce our sGRU with modification on classical GRU by adding a preservation scheme, which can keep the current time step information in a sequence to arbitrary future time step.
%an additional skip connection from an arbitrary previous unit, to adapt the cells to the additional local storyline connection in the complex structure of our photo stream input.

\subsection{The GRU Model}

The GRU, as proposed in \cite{GRU}, is a hidden unit used in RNN model for capturing long-range dependencies in sequence modeling. It becomes more and more popular in sequence modeling, as has been shown to perform as well as LSTM \cite{LSTM} on sequence modeling tasks \cite{NIPSDLW2014Chung} but with much more compact structure. An RNN has multiple GRU cells, where each cell models the memory at a time of recurrence. In Figure 2(a), the circuit in black shows the graphical depiction of the GRU design. The following equations represent the operations of components:

\vspace{-5mm}
\begin{eqnarray}\label{eq:gru}
\small
\begin{split}
\hspace{30mm}&z_{t}=\sigma(W_{zx}x_{t}+W_{zh}h_{t-1})\\
\hspace{30mm}&r_{t}=\sigma(W_{rx}x_{t}+W_{rh}h_{t-1})\\
\hspace{30mm}&\widetilde{h}=tanh(W_{hx}x_t+W_{hh}r_t\odot{h_{t-1}})\\
\hspace{30mm}&h_{t}=z_{t}\widetilde{h}+(1-z_{t})h_{t-1}\\
\end{split}
\end{eqnarray}
\vspace{-4mm}

where $t$ is the current time step, $x_t$ is the input, $\widetilde{h}$ is the current hidden state and $h_{t}$ is the output. $z_{t}$ and $r_{t}$ are \emph{update gate} and \emph{reset gate}, respectively. Note that we omit the bias $b$ in every equation for compact expression. By definition, the update gate $z_{t}$ controls how much information from the previous will be delivered to the current, and the reset gate $r_{t}$ decides how much information from the pervious to drop if it is found to be irrelevant to the future. In other words, if $r_{t}$ is close to 0, the current unit is forced to ignore the previous $h_{t-1}$ and reset with the current input only.

\subsection{The sGRU Model}

At the time step $t$, the motivation of reset gate $r_{t}$ in GRU is to stop irrelevant information in $h_{t-1}$ from going any further to future \cite{GRU}. We notice that $r_{t}$ is controlled by comparing $h_{t-1}$ to the current input $x_{t}$. The logic behind this design of GRU, in other words, is that the current input $x_{t}$ decides whether or not $h_{t-1}$ can be passed through the reset gate to the future. If $x_{t}$ believes $h_{t-1}$ is irrelevant, all the information from previous units will be dropped and will not flow to future.

%As we introduced the intuition of pseudo-gap previously, we now give the formal definition to it. In a sequence network of GRU along time ${T}$, a pseudo-gap is a tuple $(p,t)$ that the gap through $(p,p+1)$ are large to trigger dropping of information, which can be used at time $t$.

However, the above logic is harmful when pseudo-gap appears. Specifically, if the pseudo-gap happens between time step $t-1$ and $t$ while $t-1$ is useful to predict at future $t+i$, the information in $h_{t-1}$ will be droped at $t$ and never passed to  $t+i$. Just as the case happens at time step $t=2$ and $i=2$ in figure 1. According to our statistics, the pseudo-gap problem exists in 74.9\% of story photo streams. Thus, the design of reset gate in GRU will lose much useful information and impair to the modeling of visual stories. Note that this issue also exists in other type of RNNs, such as LSTM since it employs a similar dropping scheme by its forget gate \cite{LSTM} to keep its efficient but limited memory, thus cannot practically learn the long-term dependency either \cite{LongTerm}.

To this end, we propose a preservation scheme to compensate the missing information at pseudo-gap. As shown in figure 2, we add the red part to the original design of GRU. If we know that the output at time step $p$ is needed at $t$, or we call the ordered pair ($p$, $t$) a \emph{skip}, indicated by skip matrix $R$ where $R_{pt}=1$. The name \emph{skip} here follows the same intuition as in figure 1 in introduction. The cell of $p$ is called \emph{skip-ancestor}, and the cell at $t$ is called \emph{skip-descendant}. As shown in figure 2, the information of $h_{p}$ has been saved for a delay time $|t-p|$ and is reused together with $h_{t-1}$ to predict the hidden of $\widetilde{h}$ at time step $t$. Thus we name the new unit \emph{skip Gated Recurrent Unit} (sGRU), and define its operations as:

\vspace{-8mm}
\begin{eqnarray}\label{eq:sgru}
\small
\begin{split}
\hspace{20mm}&z_{t}=\sigma(W_{zx}x_{t}+W_{zh}h_{t-1})\\
\hspace{20mm}&r_{t}=\sigma(W_{rx}x_{t}+W_{rh}h_{t-1})\\
\hspace{20mm}&\bm{{s_{t}=\sigma(W_{sx}x_{t}+W_{sh}h_{p})}}\\
\hspace{20mm}&\widetilde{h}=tanh(W_{hx}x_t+W_{hh}r_t\odot{h_{t-1}}+\bm{ \sum\nolimits_{p}{R_{pt}\cdot W_{hp}s_t\odot{h_{p}}}})\\
\hspace{20mm}&h_{t}=z_{t}\widetilde{h}+(1-z_{t})h_{t-1}\\
\end{split}
\end{eqnarray}
\vspace{-4mm}

where the $s_{t}$ represents the \emph{skip gate} to control how much information is used from $h_{p}$. Similar to other gates, the design of skip gate $s_{t}$ is controlled by current input and skip ancestor's output. Prior to the learning of sGRU, the skip relation matrix $R$ is calculated offline, in our case, by an unsupervised photo alignment model (Section 3.2). Intuitively, the skip relation corresponds to the local storyline structure in the story photo streams. Note that $R$ is a sparse matrix where each row contains up to one non-zero value, which equals to 1 and indicate the skip ancestor.

The advantages of such a design are two folds. First, the preservation scheme with delay can maintain the longer dependency and bridge the pseudo-gap. The information missed by general GRU at pseudo-gap will be kept in sGRU. Second, the skip gate ensures the non-linear mapping through skips. Otherwise, one may alternatively takes linear combination of $h_{p}$ with $x_{t}$ in the formula \ref{eq:sgru}, just as a common variations on LSTM \cite{CLSTM,gLSTM}. However, we found the linear combination empirically yield worse performance in experiments. Without the power of non-linear mapping in skip gate, these models are unable to capture the deep semantic relation between the additional skip information and the current input. Thus, our proposed sGRU model is capable of capturing the longer dependency in cross-skipping structure.

\begin{figure}[t]
\centering
   \includegraphics[width = 0.9\linewidth]{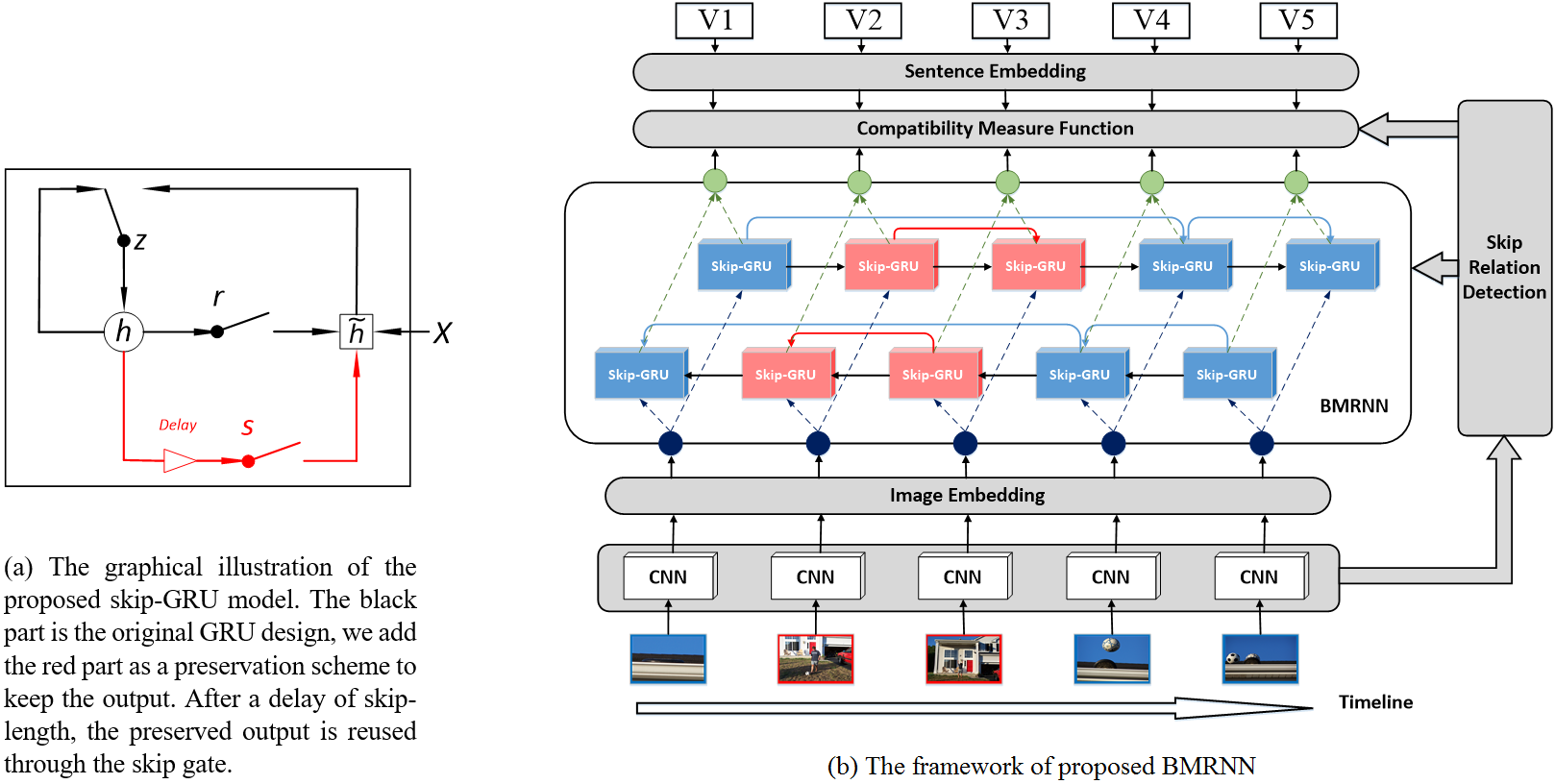}
   \caption{\small (a) Our skip-GRU model and (b) the framework of our BMRNN model.}
   \label{fig:exe}
\end{figure}
\vspace{-4mm}

\section{Architecture}

We propose a Bidirectional Multi-thread RNN to model the complex visual structure of visual stories and generate narrative language. As shown in figure 2(b), our framework includes three main parts: (1) image and sentence representation based on an pre-trained joint embedding, (2) a skip relation detection and (3) a BMRNN network integrated with sGRU for visual structure modeling and text story generation.

Given a photo stream denoted as $S=\{I_1,I_2,...,I_N\}$, we first extract the CNN features $C=\{fc_1,fc_2,...,fc_N\}$ from the fc7 layer of VGGNet \cite{VGGNet}. $N$ is the length of the photo stream. We then project the image features to an embedding space to product the embedding vector $X=\{x_1,x_2,...,x_N\}$. The embedding of images and sentences are jointly learned with image-sentence pairs in dataset as in \cite{OrderEmbeding}. We employ this embedding model since it is rank-preserving in learning of embedding space, which is desired in our retrieval task. Meanwhile, the CNN features of photo streams are also used to detect the skip relation $R$, Finally, the proposed BMRNN model integrates the sGRU and use the skip relation to model the complex visual flow based on input of image embedding vectors $X=\{x_1,x_2,...,x_N\}$. The sentence embedding vectors $H=\{h_1,h_2,...,h_N\}$ are predicted for retrieval. We introduce the skip relation detection and BMRNN model in Section 3.1 and Section 3.2, respectively, followed by the the loss function and training process specified in Section 3.4.

\subsection{Skip-Relation Detection}

To detect the visual structure and local connections corresponding to storylines, we employ an unsupervised skip-detection based on the CNN feature of images $C=\{fc_1,fc_2,...,fc_N\}$. First, we utilize the affinity propagation (AP) \cite{AP} to cluster the images, to detect different scenes that appear in the story. We take use of AP rather than other methods because it does not explicitly need cluster numbers to be decided in advance. As many previous works have validated, CNN features perform very reasonable results in object detection and scene representation. Thus, we define the likelihood of two images sharing similar scene as their inner product of CNN features:

\vspace{-8mm}
\begin{eqnarray}\label{eq:skipmatrix}
\small
\begin{split}
\hspace{30mm}&sim(I_{i},I_{j})=fc_{i}*fc_{j}^T\\
\end{split}
\end{eqnarray}
\vspace{-8mm}

Then, with the clustering results, we connect the images in same cluster along the original temporal order. The skip relation matrix $R$ can be produced where at position $(i,j)$:

\vspace{-6mm}
\begin{eqnarray}
\small
\hspace{25mm}&R_{ij}=
\left\{
             \begin{array}{lr}
             1 & skip (i,j) exist  \\
             0 & otherwise
             \end{array}
\right..
\end{eqnarray}
\vspace{-6mm}

\subsection{BMRNN Model}

The role of BMRNN is to model the complex structure of visual stories. In our problem, the bidirectional framework can consider moments of both past and future simultaneously. The sGRU is used as basic unit, whose additional skip connections lead to the multi-thread form of BMRNN. We rewrite the sGRU in the formula \ref{eq:sgru} into a compact form: $(z_t,r_t,s_t,\widetilde{h},h_t) = sGRU(x_t,h_{t-1},R,h_p;\bm{W})$, to define the operations of our BMRNN components as:

\vspace{-5mm}
\begin{eqnarray}\label{eq:BMRNN}
\small
\begin{aligned}
\hspace{20mm}(z_t^f,r_t^f,s_t^f,\widetilde{h}^f,h_t^f) &= sGRU(x_t,h_{t-1}^f,R,h_p^f;\bm{W}^f)\\
(z_t^b,r_t^b,s_t^b,\widetilde{h}^b,h_t^b) &= sGRU(x_t,h_{t+1}^b,R^T,h_p^b;\bm{W}^b)\\
h_{t} &= W_h^f h_{t}^f+W_h^b h_t^b
\end{aligned}
\end{eqnarray}
\vspace{-6mm}

where the superscript $f$ indicates the forward pass and $b$ is for the backward pass. The two passes neither have any inter-communication nor share any parameters, except the input $x_t$. Each pass is learned independently and flexibly. In the test, we empirically validate that this independent parameter setting achieves better performance than the sharing one. At last, the outputs of two passes are merged to generate the final output of whole framework. In training, we learn the parameters $\bm{W}=\{\bm{W}^f,\bm{W}^b,W_h^f,W_h^b\}$ in the formula \ref{eq:BMRNN} of our model. Note that the skip relation matrix of backward pass $R^b$ can be easily obtained by transpose of the forward pass $R^f$, i.e. $R^b=(R^f)^T$.

\subsection{Training}

Previous work \cite{Kim} only measures the global compatibility between image streams and the sentence sequences, without taking into account the local compatibility in the structure of storyline. Instead, we define a new compatibility score function that consider both of them:

\vspace{-5mm}
\begin{eqnarray}\label{eq:compatibility}
\small
\begin{split}
\hspace{20mm}&c(H,V)=\alpha{\sum\nolimits_{t=1...N}{h_{t}\cdot{v_{t}}}}+(1-\alpha){\sum_{i}{m(H_{i},V_{i})}}\\
where\hspace{20mm}&m(H_{i},V_{i})=\frac{1}{n_{i}}\sum\nolimits_{h\in{H_{i}},v\in{V_{i}}}{h\cdot{v}}
\end{split}
\end{eqnarray}
\vspace{-2mm}

Where $H=\{h_1,h_2,...,H_N\}$ is the output of our BMRNN model with a story photo stream input and $V=\{v_1,v_2,...,v_N\}$ is the sentence sequence to be matched. $(H_{i},V_{i})\subset {(H,V)}$ denotes a sub-story with length $n_{i}$, and $m(H_{i},V_{i})$ calculates the local compatibility. For the cost function, we employ the contrastive loss with margin as in \cite{Unifying} \cite{Kim} \cite{OrderEmbeding}, which calculates:

\vspace{-5mm}
\begin{eqnarray}\label{eq:loss}
\small
\begin{split}
\hspace{5mm}&\sum\nolimits_{(H,V)}{\left(\sum\nolimits_{V'}{\max\nolimits{\{ 0,\gamma-c(H,V)+c(H,V')\}}}+\sum\nolimits_{H'}{\max{\{ 0,\gamma-c(H,V)+c(H',V)\}}}\right)}\\
\end{split}
\end{eqnarray}
\vspace{-5mm}

The negative (contrastive) pairs are sampled randomly from the training dataset. In our training, 127 negative pairs are generated for each positive sample. Minimization on the cost function will encourage higher compatibility between aligned image-sentence sequence pairs, and punish on random-aligned pairs. In the dataset of NYC and Disney, the negative samples are very unlikely to have the same length with the positive samples \cite{Kim}. As the result, the compatibility in the negative pairs is almost always low. So the punishment term on the negative samples do NOT take much effect in helping the model learning. However, in the SIND, all stories have the same length $N=5$. Thus the punishment term for negative samples becomes as important as that of positive term in cost function.

\section{Experiment}

In experiments, we compare our approach with a group of state-of-art methods from both exist models and variations of our own model. We evaluate the performance in two kind of measures, a series of quantitative measures and user study.

\subsection{Experiment Setting}

\subsubsection*{Dataset}

To evaluate our BMRNN model on the visual storytelling task, we use three different recently proposed dataset, the SIND \cite{VisualStorytelling}, NYC and Disneyland dataset \cite{Kim}. All the three datasets consist of sequential image-stream-to-sentence-sequence pairs. The SIND is the first dataset particularly for sequential vision-to-language task. It contains 48,043 stories with 210,819 unique photos. The image streams are extracted from Flickr and the text stories are written by AMT. Each story consists of 5 images and 5 corresponding sentences written in sense of story. The dataset has been split by the authors into 38,386 stories as training set, 4,837 as test set and 4,820 as validation set.

The NYC and Disney datasets are automatically generated from blog posts searched with travel topics “nyc” and “disneyland”, respectively, where each blog contains a stream of images and its descriptive blog post. For one image, the associated paragraph in blog post is processed to summarize a descriptive sentence. Although these sentences are not created diligently for storytelling purpose, they can be treated as a semantically meaningful authoring to the image streams \cite{Kim} under context of the travel blog posts. Thus they can also be used as qualified story datasets for our problem. Unlike SIND, the stream length is not uniform, varying from minimum length of 2 to maximum of 226 photos. We also follow the splitting of dataset in \cite{Kim} that 80\% as training set, 10\% as validation set and the others as test set.

\subsubsection*{Tasks}

We compare our approach with many state-of-art candidate methods in task of generating story sentences for photo streams. We use the image streams in test set as input to the framework and the corresponding text story as groundtruth. The models retrieve the best stories from the training set. We use the Recall@K metric and median rank to evaluate the retrieved stories. The recall@K indicate the recall rate of the groundtruth retrieval given top K candidates, and the median rank is the median rank value of the retrieved groundtruth. The higher recall@K and lower median rank value, the better performance is indicated.

\subsubsection*{Baselines}

In our experiment, we consider both state-of-art methods from the existing models and variation of our own BMRNN model. Since the visual storytelling is a promising while new research direction, there are few existed research works that have focused on the task and addressed this problem. To the best of our knowledge, the only closely related work is \cite{Kim}, a sentence sequence generation from image streams based on coherent recurrent convolutional networks (CRCN). Besides, we also take other state-of-art models in vision-to-language tasks as baseline, such as video description using CNN and BRNN (CNN+BRNN) of \cite{VideoBRNN}. Particularly, the (CNN+BRNN) model without sGRU is also a variation to our model. Thus comparing between us and (CNN+BRNN) can evaluate the effectiveness of our proposed sGRU architecture. To validate our claim on the non-linear mapping power of sGRU, we also compare to CLSTM unit where the variation is in linear fashion \cite{CLSTM}, in a bidirectional framework as us. Thus we call this baseline as (B-CLSTM) We also test the K-nearest search (1NN) and random retrieval scheme as simple baselines.

%To provide wider scope of information in sense of comparison, we directly take the reported results of baseline methods in \cite{Kim}, including (CNN+LSTM), (CNN+RNN)

\subsection{Quantitative Results and Discussion}

The quantitative results of story sentences retrieval are shown in table 1 and 2. We observe that we perform better in a large margin than other candidates baselines on all datasets, that confirms our analysis on the unique cross-skipping structure of visual stories. It can be further validated that our BMRNN model is effective to capture and leverage this skip relations to improve in visual storytelling. Specifically, comparing with (CRCN) shows that visual modeling on photo stream can better approach the visual storytelling task, rather than only capture coherence in textual domain, although they adopt an additional K-nearest candidates search to largely reduce the search scope on test set prior to retrieval. We found that we outperform (CNN+BRNN) method, which verifies the importance of our proposed sGRU architecture in capturing the cross-skipping structure. More deeply, it is because of the non-linear design that empowers the sGRU to capture the skipping information, since it achieves higher results than (B-CLSTM) where a linear scheme is used.  Note that we take use the same VGGNet fc7 feature \cite{VGGNet} in all baselines for fair comparison.

We also find out that the visual models (CNN+BRNN) and (B-CLSTM) all yield fairly higher results than (CRCN) in the retrieval metrics. This verify again our believe that modeling from visual domain than textual domain can better approach the storytelling task. The results of (B-CLSTM) is almost equal to (CNN+BRNN), showing failure of that the \emph{context} design. Since it is unable to discover a common context shared by all photos in a visual story. The (1NN) baseline shows unsatisfactory results, indicating that the visual story is never simple concatenation of individual image captions.

\begin{table}[!h]
\begin{center}
\small
\label{tab:nyc}
\begin{tabular}{|c||c|c|c|c||c|c|c|c|}
  \hline
  & \multicolumn{4}{|c|}{\textbf{NYC}} & \multicolumn{4}{|c|}{\textbf{Disneyland}}  \\
  \cline{2-9}
   &  \textbf{R@1} &  \textbf{R@5} &  \textbf{R@10} &  \textbf{Medr} &  \textbf{R@1} &  \textbf{R@5} &  \textbf{R@10} &  \textbf{Medr}\\
  \hline
  Random & 0.17 & 0.25 & 0.59 & 763 & 0.26 & 1.17 & 1.95 & 332 \\
  1NN & 5.95& 13.57 & 20.71 & 63.5 & 9.18 & 19.05 & 27.21 & 45\\
  CNN+BRNN\cite{VideoBRNN} & 16.23 & 28.7 & 39.53 & 19 & 19.97 & 37.48 & 46.04 & 14 \\
  B-CLSTM\cite{CLSTM} & 15.10 & 29.91 & 41.07 & 18 & 19.77 & 38.92 & 45.20 & 14 \\
  CRCN\cite{Kim} & 11.67 & 31.19 & 43.57& 14 & 14.29 & 31.29 & 43.2 & 16 \\
  \hline
  BMRNN & \textbf{31.73} & \textbf{46.86} & \textbf{53.50} & \textbf{8}& \textbf{37.10} & \textbf{50.71} & \textbf{58.63} & \textbf{5}\\
  \hline
\end{tabular}
\caption{\small Retrieval metrics evaluation of sentence generation on NYC and Disney dataset}
\end{center}
\end{table}

\begin{minipage}[t]{.5\textwidth}
\centering
\small
  \begin{tabular}{|c||c|c|c|c|}
  \hline
  & \multicolumn{4}{|c|}{\textbf{SIND}}  \\
  \cline{2-5}
   &  \textbf{R@1} &  \textbf{R@5} &  \textbf{R@10} &  \textbf{Medr} \\
  \hline
  Random & 0.0 & 0.04 & 0.10 & 2753  \\
  1NN & 4.8 & 13.00 & 21.07 & 74 \\
  CNN+BRNN\cite{VideoBRNN} & 21.39 & 38.72 & 46.96 & 14 \\
  B-CLSTM\cite{CLSTM} & 21.47 & 37.30 & 47.39 & 18 \\
  CRCN\cite{Kim} & 9.87 & 28.74 & 39.51 & 21 \\
  \hline
  BMRNN & \textbf{25.56} & \textbf{45.17} & \textbf{53.69} & \textbf{8}\\
  \hline
\end{tabular}
\tabcaption{\small Evaluation of sentence generation on SIND dataset.}
\end{minipage}
\hspace{7mm}
\begin{minipage}[t]{.4\textwidth}
\centering
\small
\begin{tabular}{|c|c c c|}
  \hline
   &  GT &  CRCN &  BMRNN \\
  \hline
  GT & - & 97.0\% & 86.0\% \\
  CRCN & 2.0\% & - & 13.5\%  \\
  BMRNN & 7.0\% & 67.0\% & -  \\
  \hline
  \textbf{Mean Score} & 8.48 & 3.77 & 5.19  \\
  \hline
\end{tabular}
\tabcaption{\small The evaluation results of user study. The row 1-3 are pairwise preference and the last row is the mean evaluation scores.}
\end{minipage}

\subsection{User Study}
We perform an user study to test the preference on the generated stories by groundtruth, our model and baselines. Since only (CRCN) is originally proposed for the sequential vision-to-language task, we choose this method as our baseline in our user study. We randomly sampled 200 stories from the test set of SIND dataset, each associated with three stories: story from groundtruth (GT),story generated by \cite{Kim} (CRCN) and story by our proposed method (BMRNN). Please see the figure 3 for examples. Then 10 users are invited to score on the stories with a subjective score 1-10 (Good story = 10). All the the three story, including the groundtruth stories, are read by users and evaluated.

The table 3 shows our user study results, where the last row is the mean score over all samples. We infer the user preference between two stories by comparing their scores from a same person. Equal scores indicate no preference to any method and thus are not considered in preference inference. The rows 1-3 in table 3 show pairwise preference of each method against others.

We observe that the stories by our method is better than (CRCN) in user study. Over all users, our mean score are higher than (CRCN). For particular each user, in 68.0\% cases we are more preferred than (CRCN), while in only 13.5\% cases it is opposite. It is also true in another aspects, where we are more likely to catch up groundtruth at 7.0\% than (CRCN) at 2.0\%. These results all validated that our results are more preferred than the baseline.

\section{Conclusion}

In this paper, we focus on visual storytelling task, which enables generating human-level narrative language from a photo stream. To solve the cross-skipping problem in typical stories, we propose a novel BMRNN model with sGRU architecture that is able to model the complex structure in photo streams. Experiments show the effectiveness of our model in both quantitative evaluations and a user study. The proposed BMRNN model can outperform the-state-of-art methods in vision-to-language tasks. In future, we will continue exploring the modeling of visual sequences and its related applications.

\begin{figure}[t]
   \includegraphics[width = 1.00\linewidth]{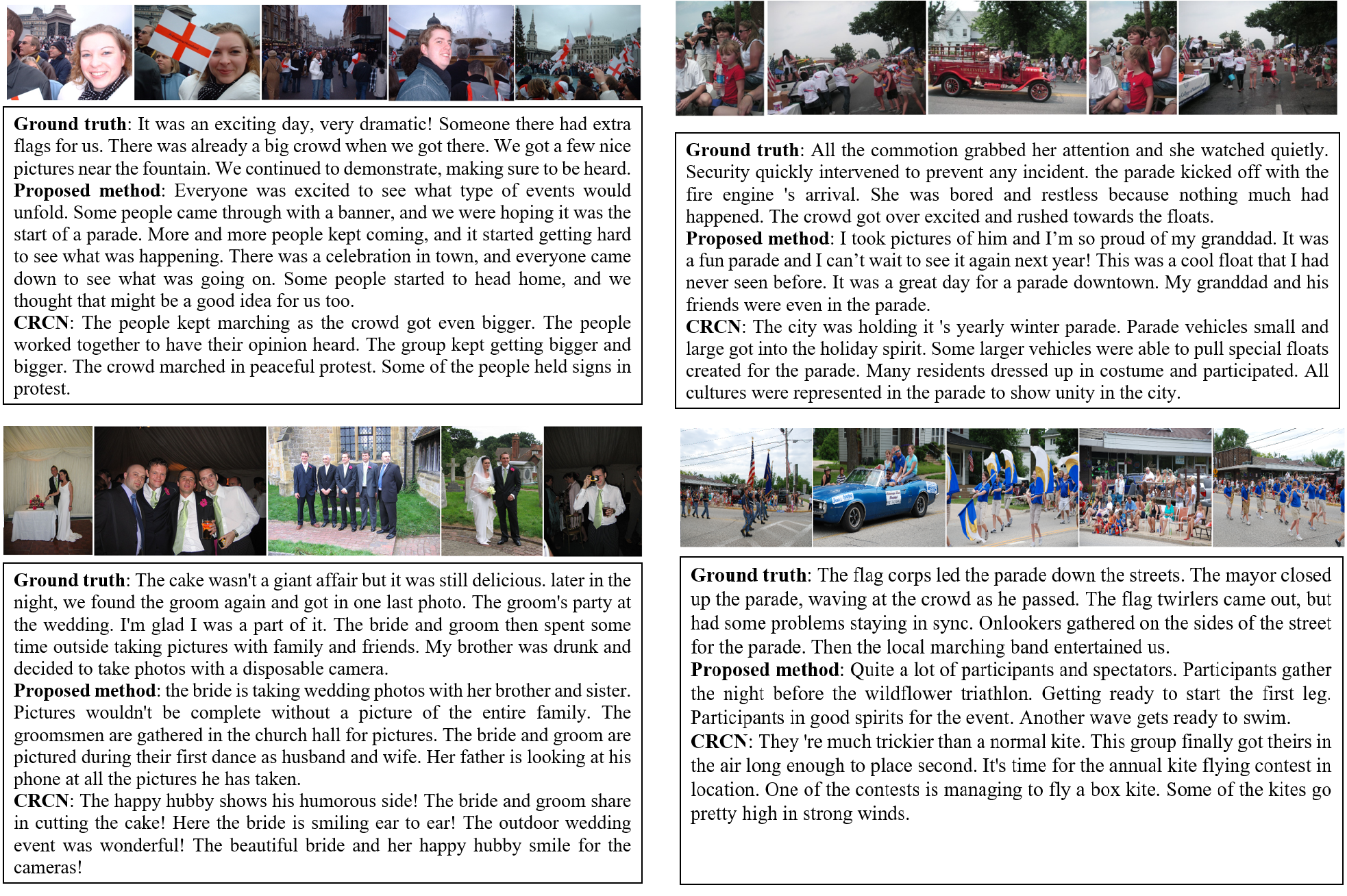}
   \caption{\small Examples of visual storytelling result on SIND. Three stories are generated for each story photo stream: groundtruth story, story generated by proposed BMRNN method and story from baseline CRCN method.}
   \label{fig:big}
\end{figure}

\newpage

{\small
\vspace{-2mm}
\bibliography{nips_2016}
\bibliographystyle{abbrvnat}
}

\end{document}